\documentclass[letterpaper, 10 pt, conference]{ieeeconf}
\usepackage[cmex10]{amsmath}
\usepackage{array}
\usepackage{url}
\usepackage[utf8]{inputenc}  
\usepackage{citesort}
\usepackage[T1]{fontenc} 
\usepackage{mathtools}
\usepackage{amssymb,amsfonts,amsmath}
\usepackage{dsfont}
\usepackage{hyperref}
\usepackage{dsfont}
\usepackage{stmaryrd}
\usepackage{bm}
\usepackage{flushend}
\usepackage{algorithm}      % microtypography

\def\reals{\mathbb{R}}
\def\(({\left(}
\def\)){\right)}                       
\def\[[{\left[}
\def\]]{\right]}

\title{\LARGE  Phase transitions and optimal algorithms \\ in high-dimensional Gaussian mixture clustering}

\author{Thibault Lesieur $^{\dagger}$, Caterina De Bacco$^{\ddagger}$ , Jess Banks$^{\ddagger}$, Florent Krzakala$^*$, Cris Moore$^{\ddagger}$ and Lenka Zdeborov\'a$\dagger$\\
$\dagger$ Institut de Physique Th\'eorique, CNRS, CEA, Universit\'e Paris-Saclay,  F-91191, Gif-sur-Yvette, France\\
$\ddagger$ Santa Fe Institute, 1399 Hyde Park Road Santa Fe, New Mexico 87501, USA \\
$*$ Laboratoire de Physique Statistique, CNRS, PSL Universit\'es \&
Ecole Normale Sup\'erieure, \\Sorbonne Universit\'es et Universit\'e Pierre \& Marie Curie, 75005, Paris, France.}

\begin{document}
\maketitle
\begin{abstract} We consider the problem of Gaussian mixture clustering in the high-dimensional limit where the data consists of $m$ points in $n$ dimensions, $n,m \rightarrow \infty$ and $\alpha = m/n$ stays finite.  Using exact but non-rigorous methods from statistical physics, we determine the critical value of $\alpha$ and the distance between the clusters at which it becomes information-theoretically possible to reconstruct the membership into clusters better than chance. We also determine the accuracy achievable by the Bayes-optimal estimation algorithm.  In particular, we find that when the number of clusters is sufficiently large, $r > 4 + 2 \sqrt{\alpha}$, there is a gap between the threshold for information-theoretically optimal performance and the threshold at which known algorithms succeed.  \end{abstract}

Clustering $m$ points in $n$-dimensional space is a ubiquitous problem in statistical inference and data science.  It is especially challenging in the (practically relevant) context of high-dimensional statistics when the dimension $n$ is large (so that there are many degrees of freedom) but the number of points $m$ is only linear in $n$ (so that the available information is limited).  An important special case is the problem of clustering points generated by Gaussian mixture models (GMM). These are probabilistic models where all the data points are generated from a mixture of a {\it finite} number $r$ of Gaussian distributions.

In this paper, we are interested in understanding the fundamental limits on our ability to cluster points generated from GMM in the high-dimensional regime, when $m,n \to \infty$ while $m/n=\alpha$ is finite. We consider both a {\it computational} and an {\it information theoretic} viewpoint, and wish to answer the following questions: Given data generated by a GMM, (i) what is the best possible estimate, information-theoretically, of the parameters of the Gaussian mixtures, and of the individual assignments of the points? (ii) Are such optimal estimates computationally feasible in practice by polynomial-time algorithms? (iii) How do standard, widely used, methods such as principal component analysis \cite{hoyle2004principal,10.2307/3481698} compare with these optimal predictions?

We adress these questions by taking advantage of the recent burst of activity in the related problem of low-rank matrix factorization, using the cavity method from statistical physics \cite{mezard2009information} and the associated approximate message passing (AMP) algorithm \cite{rangan2012iterative,NIPS2013_5074}.  We shall not attempt at mathematical rigor here, but it is worth noting that recent progress on closely related problems \cite{javanmard2013state,krzakala2016mutual,DBLP:journals/corr/BarbierDMKLZ16} is likely transferable to the present situation.
%many (but not all) of the stated results could be readily proven with existing methods, a task left open for future work.

\section{Model and setting} 
Consider data generated by a GMM with $r$ clusters: for each $k \in \{1,\ldots,r\}$ we draw each coordinate of $V_k \in \reals^n$ from a Gaussian of zero mean and unit variance. We then generate $m\!=\!\alpha n$ points $x_1,\ldots,x_m$ $\in \reals^n$ independently as follows: for each $j$, choose $t_j \in \{1,\ldots,r\}$ uniformly at random, and set $x_j = \sqrt{\rho/n}\, V_{t_j} + U_j$ where $U_j\in \reals^n$ and has mean $0$ and variance $\Delta$ in each coordinate.  Here $\rho$ is a parameter playing the role of the signal to noise ratio and the $\sqrt{\rho/n}$ factor ensures that the displacement of the centers is of the same order as the fluctuation of a high-dimensional Gaussian point around the surface of a sphere of radius $n$. We consider the limit $n\to \infty$, while $\alpha,\rho,\Delta=O(1)$. 
%the squared distance of each centroid from the origin is $\rho \pm O(1/\sqrt{n})$. $\Delta$ describes the spread of each of the Gaussians. Both $\rho,\Delta = O(1)$. This factor $\sqrt{\rho/n}$ ensures that the problem is neither trivially hard nor trivially easy. For simplicity we assume that the clusters are isotropic.  

Using this generative model makes the clustering problem a non-trivial one: the clusters overlap significantly; assigning points to clusters without errors is impossible and it is also impossible to recover exactly the parameters of the GMM.  We thus ask instead what is the best possible overlap with the ground-truth clusters.  It is instrumental to view this problem as noisy low-rank matrix factorization.  Specifically, let $V$ denote the $n\!\times\!r$ matrix whose $k$th column is $V_k$, and let $S$ be the $m\! \times\!r$ matrix where $S_{jk}\!=\!1$ if $t_j\!=\!k$ and $0$ otherwise. We denote by $v_i, s_j \in \reals^r$ the $i$th and $j$th rows of $V$ and $S$; note that $v_i$ is the vector of the $i$th coordinate of the cluster centroids.
%whose $j$th row has a $1$   (where each element of $v_k$ is Gaussian with variance $1$ and where we have recentered the $v_k$ such that the average of the $r$ vectors is exactly $0_{\mathbb{R}^n}$). We then generate $m = \alpha n$ points $x_1,\ldots,x_m \in \R^n$ in the following way: for each $1 \le j \le m$, we choose $t(j)$ uniformly in $[1,~k]$ and denote $\bf S$ the $m \times k$ matrix whose lines are $S_{jk}=\delta_{k,t(j)}$. 
Then the observed data $X$ is an $n \times m$ matrix
\begin{equation}
	X = \sqrt{\frac{\rho}{n}} V S^\top + U \, ,
\label{model:gauss}
\end{equation}
where $U$ is a Gaussian i.i.d. $n \times m$ random matrix with mean $0$ and variance $\Delta$.  Thus $X$ is a noisy observation of a matrix $V S^\top$ of rank $r$. 
%We are using generic $\Delta$ for the discussion of this section, but 
We shall use without loss of generality $\Delta=1$, as other values of $\Delta$ simply cause a rescaling of the signal-to-noise ratio $\rho$. 

Assuming the data is generated by model (\ref{model:gauss}), we want to know how well we can reconstruct the true cluster assignments $\{t_j\}$ and cluster centers $\{V_k\}$.   To answer these questions, we use Bayesian inference. 
The generative model for the data is
\begin{align*}
&P(X,S,V)
=P(S) P(V) \,P(X|S,V) \\
&= \prod_{i=1}^n P_v(v_i) 
\prod_{j=1}^m P_s(s_j)
\prod_{i=1}^n \prod_{j=1}^m e^{-\frac{1}{2\Delta}\left(X_{ij} - \frac{\sqrt{\rho} v_i^\top s_j}{\sqrt{n}}\right)^2} \,,
\end{align*}
%The fact that this random process factorises over the line of $V$ and $S$ is part of the description of the model.
where $P_v$ and $P_s$ are the probability distributions from which the cluster centers $v_i$ and assignments $s_j$ are drawn. We take $P_v$ to be Gaussian with mean zero and covariance matrix $I_r$.  Since the labels $t_j$ are uniform, $P_s$ is the uniform distribution over the $r$ canonical basis vectors
\begin{eqnarray}
P_v(v) &=& \frac{1}{(2 \pi)^{r/2}}\exp\left(\frac{-\Vert v \Vert_2^2}{2}\right) \,, \label{Pv}
\\
P_s(s) &=& \frac{1}{r}\sum\limits_{i = 1 \cdots r} \delta\left( s - \vec{e_i} \right) \,. \label{Ps}
\end{eqnarray}
Using Bayes' rule to compute the posterior probability of $S$ and $V$ given the data $X$, one gets
\begin{align}
 P(S,V|X) &= \frac{1}{P(X)}  \prod_{i=1}^n P_v(v_i) \prod_{j=1}^m P_s(s_j)  \nonumber \\
&e^{-\frac{1}{2 \Delta} \sum_{ij} \left( X_{ij} - \frac{\sqrt{\rho} v_i^\top s_j}{\sqrt{n}} \right)^2 }\,.\label{DensityProb}
\end{align}

We denote the ground truth cluster centers and assignments used to generate $X$ as $V_0$ and $S_0$. We assume that the correct prior distributions~\eqref{Pv} and~\eqref{Ps} are known to the algorithm and analysis; specifically, the correct number of clusters $r$ and the parameters $\rho, \Delta$. We call this the {\it Bayes-optimal} setting.  The question of what happens when the model is incorrect is also interesting, but in this article we let is aside. We note, however, that taking the limit $\rho\! \to\! \infty$ and $\Delta \!\to\!  0$ in the Bayesian setting would correspond to the minimization problem  that the $k$-means algorithm~\cite{lloyd1982least}  is trying to solve. The objective of {\it soft} $k$-means corresponds instead to the Bayesian setting in the limit $\rho\!\to\! 0$ while retaining a nonzero noise level $\Delta$.  Spectral methods, on the other hand, are equivalent to Bayesian inference where one abandons the hard constraint that the labels $t_j$ are discrete, and effectively replace $P_s$ with a Gaussian prior.

The above Bayesian approach converts the problem of estimating an assignment of points to clusters that maximizes the number of correctly assigned points to the problem of computing marginal probabilities of the posterior distribution. The mean squared error in estimating $V_0$ from $X$ is minimized by using the (marginalized) conditional expectation of $V$ given the data~\cite{cover2012elements}. For estimation of $S_0$ we aim to minimize the number of mis-classifications. The corresponding Bayes-optimal estimator $\hat S_{\rm MaxProb}$ is constructed by assigning each data point to the cluster that, according to the marginals of (\ref{DensityProb}), it is most likely to belong to. Given $\hat S_{\rm MaxProb}$, the \emph{overlap} between the estimated cluster assignments and the ground truth is defined as
\begin{eqnarray}
{\rm ErrorRate} &=& \frac{1}{m}{\rm Tr}\left( \hat S_{\rm MaxProb}^\top  S_0 \right)
\\
{\rm Overlap} &=& \frac{{\rm ErrorRate} - 1/r}{1 - 1/r} \,. \label{eq:overlap}
\end{eqnarray}
The ${\rm ErrorRate}$ is the percentage of correctly assigned points. This rate is 1 when the reconstruction is perfect. When taking the assignment at random the ${\rm ErrorRate}$ is $1/r$. The normalization in (\ref{eq:overlap}) is introduced in order to have a quantity that varies between $0$ meaning no better reconstruction that chance, to $1$ meaning perfect reconstruction.  

The marginal posterior probabilities of the $v_i$ and $s_j$ are, however, difficult to compute in general. This is because the interaction terms involving the observations $X_{ij}$ couple these variables together.
% and prevent the distribution (\ref{DensityProb}) from factorizing into a product of marginal distributions. 
%One option, unfortunately a slow one, is to try to compute the marginals by using Monte Carlo to sample directly from the posterior distribution. Instead, 
This is where we turn to techniques from the statistical physics of disordered systems which allow marginalization of joint distributions such as~\eqref{DensityProb}. 

\section{Relation to previous works} 
Study of the Bayes-optimal estimation in Gaussian mixture clustering in the regime of finite $\alpha$ is restricted to the statistical physics literature. Results is statistic are either for $\alpha \to \infty$ or for algorithms (such as spectral ones) that are suboptimal in the present setting. Existing literature that treats the Bayes-optimal setting either considered the problem only algorithmically \cite{NIPS2013_5074} or analyzes it only for two clusters, $r=2$, \cite{watkin1994optimal,barkai1994statistical,biehl1994statistical,buhot1998phase}. The main contribution of this paper is to extend the analysis to the general (finite) number of clusters and realize that the case with several clusters behaves considerably differently from $r=2$. 

Our approach relies on Approximate Message Passing (AMP) algorithm and its theoretical analysis, a large part of which was born in the statistical physics, along with the cavity and replica approach \cite{MezardParisi87b,mezard2009information,zdeborova2015statistical}. In the present case, the AMP algorithm is very close to the TAP equations \cite{thouless1977solution}. AMP for Gaussian mixture clustering is a case of AMP for low-rank matrix factorization as  has been written in \cite{rangan2012iterative,NIPS2013_5074}. State evolution (SE) \cite{bayati2011dynamics,javanmard2013state,bayati2015universality} is a theoretical technique, closely related to the cavity method \cite{MezardParisi87b}, that allows to exactly characterize the behavior of the AMP algorithm in the limit of infinite system size. A number of works on low-rank matrix factorization used recently this approach \cite{DBLP:journals/corr/DeshpandeM14,deshpande2015asymptotic,lesieur2015mmse}. The AMP, the state evolution and associated Bethe free energy were presented recently in a rather generic form in \cite{lesieur2015mmse} and we use these results extensively here in the specific case of GMM clustering and analyze the associated phase transitions and phase diagrams.

Proving the results presented here in full generality, for a general case of low-rank matrix factorizations, is a task that has attracted a lot of attention recently. In particular the replica prediction in the symmetric rank-$1$ case has been proven rigorously in a number of situations \cite{KoradaMacris,DBLP:journals/corr/DeshpandeM14,deshpande2015asymptotic,krzakala2016mutual} and a very generic proof now exists \cite{DBLP:journals/corr/BarbierDMKLZ16}. This strongly strengthens the claim that the replica predictions for the non-symmetric matrices (which we use here) are exact, despite the conjecture being still open.

\section{Summary of Main Results}
\label{mainresults}
By analyzing the Bayes-optimal estimation in the Gaussian mixture clustering we find that depending on the values of $\rho$, $r$ and $\alpha$ (without loss of generality we consider from now on $\Delta=1$), the problem appears in one of the following three phases:
\begin{itemize}
\item {\bf EASY:} The theoretically optimal reconstruction is better than chance both for the clustering problem and for finding the centroids and the AMP algorithm described in this article is able to reach this performance in the limit of large systems.
\item {\bf IMPOSSIBLE:} The theoretically optimal reconstruction does not perform better than chance. The matrix $X$ contains no exploitable information on the assignment of each point.  For the centroids the best estimate is given by taking the mean of all the data points. No algorithm can exist to recover the assignment of the points better than chance.
\item {\bf HARD:} The theoretically optimal reconstruction gives a result that is better than chance both for the clustering problem and for finding the centroids. {\it However}, without using prior knowledge on the assignment of the points or on the centroids, the AMP algorithm is not able to converge toward its optimal fixed point. In fact, the AMP fixed point reached from an uninformed initialization does not provide any information on the assignment of the points. From the current knowledge, it is plausible that this phase is computationally hard not only for AMP but for all polynomial algorithms. 
\end{itemize}

Depending on the number of clusters (that is, the rank~$r$) and the sample-to-dimension ratio $\alpha$ we identified a boundary between two regimes at
\begin{equation}
 \label{CriteriaFirstOrder}
r_c = 4 + 2 \sqrt{\alpha} \, .
\end{equation}

\begin{itemize}
\item  If the number of clusters is small enough, $r < r_c$ there are only two phases in the problem as the signal-to-noise parameter $\rho$ increases: The impossible and the easy one. This among others means that in the large size limit we are always able to reach Bayesian optimal reconstruction performance using the AMP algorithm. These two phases are separated by a sharp impossible/easy phase transition when
\begin{equation}
\rho_c = \frac{r}{\sqrt{\alpha}}   \, .
\end{equation}
The problem is easy when $\rho > \rho_c$ and impossible otherwise.
\item  If instead the number of clusters is large enough, $ r > r_c$ the three phases above are observed when decreasing $\rho$.
\begin{itemize}
\item When $\rho > \rho_c=r/\sqrt{\alpha}$ we are in the {\bf EASY} phase.
\item When $\rho_c > \rho > \rho_{\rm IT}$ we are in the {\bf HARD} phase.
\item When $\rho_{\rm IT} > \rho $ we are in the  {\bf IMPOSSIBLE} phase. 
\end{itemize}
IT here stands for information theoretic. 
We also compute the asymptotic behavior of $\rho_{\rm IT}$ for a large number of clusters and found that
\begin{equation}
\rho_{\rm IT}(r,\alpha) = 2\sqrt{\frac{r \log r}{\alpha}}(1 + o_r(1))\,,
\end{equation}
\end{itemize}
This means that for $r$ large the HARD phase is very broad. Note that this last asymptotic result has been recently proven rigorously in \cite{banks2016information} using the first and second moment methods, another indication of the correctness of the cavity/replica assumptions. 

In what follows, we give statistical physics justifications for these claims, based on the analysis of AMP and on the interpretation of the Bethe free energy (related to the mutual information) as the exact one. We hope this will motivate further rigorous studies in this direction.

\section{Approximate Message Passing (AMP)}
We recall the AMP algorithm here using the notations of \cite{lesieur2015mmse}.  Let us define two {\it denoising} functions ${f_v}(A,B) \in \mathbb{R}^{r \times 1}$ and ${f_s}(A,B) \in \mathbb{R}^{r \times 1}$ as the means of the probability distributions \begin{eqnarray}
 \frac{1}{{\cal Z}_v(A,B)} \,P_v(v) \exp\!\left(B^\top v - \frac{v^\top A v}{2} \right) \label{F_in_v} \,,
\\
 \frac{1}{{\cal Z}_s(A,B)} \,P_s(s) \exp\!\left(B^\top s - \frac{s^\top A s}{2} \right) \label{F_in_s} \,,
\end{eqnarray}
where ${\cal Z}_v(A,B)$ and ${\cal Z}_s(A,B)$ are normalization factors. Here $B$ is an $r$ dimensional column vector and $A$ is an $r \times r$ matrix.  For GMM clustering priors (\ref{Pv}-\ref{Ps}), we thus find
\begin{eqnarray}
f_v(A,B) &=& (I_r + A)^{-1} B \, ,  \label{fv_GMM}
\\
{f_{s ,k}}(A,B) &=& \frac{\exp(B_k - A_{kk}/2)}{ \sum\limits_{l=1, \cdots, r} \exp(B_l - A_{ll}/2)} \, . \label{fs_GMM}
\end{eqnarray}
Taking the derivative of $f_v(A,B)$ and $f_s(A,B)$ with respect to $B$ yields the covariance matrices of (\ref{F_in_v}) and (\ref{F_in_s}), and we thus define
\begin{eqnarray}
{f'_v}(A,B) = \left( \frac{\partial f_v}{\partial B}\right) &\in& \mathbb{R}^{r \times r}\, ,
\\
{f'_s}(A,B) = \left( \frac{\partial f_s}{\partial B}\right) &\in& \mathbb{R}^{r \times r}\, .
\end{eqnarray}

\begin{algorithm}[h]
  \caption{AMP for clustering mixtures of Gaussians}
  \label{alg:ampsym}
    {\bf Input:} Data ${\bf X}$ and initial condition $\hat v^{\rm init}, \hat s^{\rm init}$.
    \begin{eqnarray}
    \forall t &\in& \{0,1\},\nonumber
    \\\hat{v}_{i}^{t} &\gets & \hat v^{\rm init}_i ,\: \: \hat{s}_{j}^t \gets \hat{s}^{\rm init}_j ,\nonumber
\\ \sigma_{v,i}^{t} &\gets& 0^{\mathbb{R}^{r \times r}},\: \: \sigma_{s,j}^{t} \gets 0^{\mathbb{R}^{r \times r}} \nonumber
\end{eqnarray}
\text{For $t\ge 1$ compute}
\footnotesize
\begin{eqnarray}
\forall i \in [1;n],\,B_{v,i}^{t} &\gets& \sqrt{\frac{\rho}{n}} \sum\limits_{j=1 \cdots m} X_{ij} \hat{s}_j^t - \frac{\alpha \rho}{m} \sum\limits_{j = 1 \cdots m} \sigma_{s,j}^t  \hat{v}_i^{t-1} \nonumber
\\
A_v^{t} &\gets&  \frac {\alpha \rho}{m} \sum\limits_{j=1 \cdots m} {\hat{s}_j^t \hat{s}_j^t}^\top \nonumber
\\
\forall i \in [1;n],\,\hat{v}_i^t &\gets& f_v\left( A_v^t,B_{v,i}^t \right),\: \: \sigma_{v,i}^t \gets {f'_v}\left( A_v^t,B_{v,i}^t \right) \nonumber
\\
\forall j \in [1;m],\,B_{s,j}^{t} &\gets& \sqrt{\frac{\rho}{n}} \sum\limits_{i=1 \cdots n} X_{ij} \hat{v}_i^t - \frac{\rho}{n} \sum\limits_{i = 1 \cdots n} \sigma_{v,i}^t  \hat{s}_j^{t} \nonumber
\\
A_v^{t} &\gets&  \frac {\rho}{n} \sum\limits_{i=1 \cdots n} {\hat{v}_i^t \hat{v}_i^t}^\top \nonumber
\\
\forall j \in [1;m],\,\hat{s}_j^{t+1} &\gets& f_s\left( A_s^t,B_{s,j}^t \right),\: \: \sigma_{s,j}^{t+1} \gets {f'_s}\left( A_s^t,B_{s,j}^t \right) \nonumber
    \end{eqnarray}
   \normalsize
Here $\forall i \in [1;n] \,B_{v,i}^t \in \mathbb{R}^{r \times 1},\, \forall j \in [1;m] \,B_{s,j}^t\in \mathbb{R}^{r \times 1}$ and $A_v^t \in \mathbb{R}^{r \times r},A_s^t\in \mathbb{R}^{r \times r}$.

Iterate till convergence.
\end{algorithm}

% FLORENT: INSTEAD OF THIS; JUST WRITE THE ONE YOU USE}
% There are multiple equivalent choice for the update order of the AMP equations. One update order of the AMP equations can lead to another update order of the State Evolution equations that allows one to analyze the AMP equations.

\addcontentsline{file}{sec_unit}{entry}
\section{State Evolution (SE)}
The SE is a method for tracking the evolution of an AMP algorithm along iterations in the large system-size limit. A physics derivation of these SE equations can be found in \cite{lesieur2015mmse}, rigorous proof follows from \cite{javanmard2013state}. We first introduce the following order parameters
\begin{eqnarray}
M_v^t &=& \frac{1}{n} \sum\limits_{i = 1\cdots n} {\hat{v}^t_i} v_{0,i}^\top \, ,
\\
M_s^t &=& \frac{1}{m} \sum\limits_{j = 1\cdots m} \hat{s}^t_j s_{0,j}^\top \, ,
\end{eqnarray}
where $\hat{v}^t_i$ $\hat{s}^t_j$ are the estimators of the posterior means of the variables $v_i$ and $s_j$ at time $t$ defined by
\begin{eqnarray}
\hat{v}^t_i &=& f_v(A_v^t,B_{v,i}^t) \,,
\\
\hat{s}^{t+1}_j &=& f_s(A_s^t,B_{s,j}^t) \,.
\end{eqnarray}

%The $\sigma_{v,i}^t$ and $\sigma_{s,j}^t$ matrices are estimators of the posterior covariances matrices at time $t$ of variables $v_i$ and $s_j$ defined by
%\begin{eqnarray}
%\sigma_{v,i}^t &=& f_v(A_v^t,B_{v,i}^t) \,,
%\\
%\sigma_{s,j}^{t+1} &=& f_s(A_s^t,B_{s,j}^t) \,.
%\end{eqnarray}

The $r\times r$ matrices $M$ describe the overlap of the estimators with the ground truth solution. 
%It follows from the tower property of the Bayesian formula that in the present (Bayes-optimal) setting $M^t_v = Q^t_v$ and $M^t_s = Q^t_s$ (These are often called the Nishimori relation \cite{zdeborova2015statistical}) so that 
The SE reads \cite{lesieur2015mmse}
%\small
\begin{eqnarray}
\label{SE_M_V}
M_v^{t} = \mathbb{E}_{W,v_0}\! \left[ f_v \! \left(\alpha \rho M_s^t , \alpha \rho M_s^t v_0 + \sqrt{\alpha \rho M_s^t} W\right)v_0^\top \right] %\triangleq g_v(\alpha \rho M_s^t)
\,,
\\
\label{SE_M_S}
M_s^{t+1} = \mathbb{E}_{W,s_0}\left[ f_s\left( \rho M_v^t , \rho M_v^t s_0 + \sqrt{\rho M_v^t} W\right)s_0^\top \right] 
%\triangleq g_s(\rho M_v^t)
\,,
\end{eqnarray}
%\normalsize
where $W$ is an $r$-dimensional Gaussian variable with zero mean and unit variance in each coordinate, and $v_0$ and $s_0$ are random variables distributed according to $P_v$ (\ref{Pv}) and $P_s$ (\ref{Ps}) respectively.

A fixed point of the SE equations is a local extremum of the so-called Bethe free energy $\phi_{\rm B}$ \cite{lesieur2015mmse}
%\small
\begin{multline}
\phi_{\rm B}(M_v,M_s) = \frac{\alpha \rho{\rm Tr}(M_v M_s^{\top})}{2}  \\ -\mathbb{E}_{W,v_0}  \left[ \log {\cal
         Z}_v(\alpha \rho M_s,\alpha \rho M_s v_0 +\sqrt{\alpha \rho M_s} W) \right] \\ - \alpha\mathbb{E}_{W,s_0}  \left[ \log {\cal
         Z}_s(\rho M_v,\rho M_v s_0+ \sqrt{\rho  M_v}W) \right] \,,
 \label{BetheFreeEnergy_UV_DE}
\end{multline}
%\normalsize
where the quantities ${\cal Z}_v$ and ${\cal Z}_s$ are the normalization factors from (\ref{F_in_v}) and (\ref{F_in_s}). 
A stable fixed point of the SE equations above is a local minimum of $\phi_{\rm B}(M_v,M_s)$ as can be seen by taking partial derivatives with respect to $M_v$ and $M_s$. 
%In particular we get the following equality on $\phi_{SV^\top}(M_v,M_s)$:
%\begin{eqnarray}
%\frac{\partial \phi_{SV^\top}(M_v,M_s)}{\partial M_v} &=& -\frac{\alpha}{2}
%\left({\rm Update}(M_s) - M_s\right)
%\left[g_s(M_v) - M_s\right]
%\,,
%\\
%\frac{\partial \phi_{SV^\top}(M_v,M_s)}{\partial M_s} &=& -\frac{\alpha}{2}
%\left({\rm Update}(M_v) - M_v\right)
%\left[g_v(M_s) - M_v\right]
%\,,
%\end{eqnarray}
%where $g_s(M_v)$ and $g_v(M_s)$ are the updated values of $M_s$ and $M_v$ respectively, as defined in \eqref{SE_M_V} and \eqref{SE_M_S}. 

It follows from the replica theory of statistical physics that
if there are multiple stable fixed points of (\ref{SE_M_V}-\ref{SE_M_S}), only the one with minimal free energy $\phi_{\rm B}$ corresponds to the performance of the Bayes-optimal estimation. On the other hand the performance reached by the AMP algorithm corresponds to the fixed point of SE with largest error. %We use this property heavily in the present paper, but the proof of this is left for future work. 

%\subsection{Performance of Principal Component Analysis (PCA)}
%The overlap achieved by PCA is also be derived using the SE equations and yields 
%\small
%\begin{eqnarray}
%&m_v = \max\left( \frac{\alpha \rho^2 - r^2}{r + \rho \alpha} , 0 \right) \label{PCA_MSE}
%\\
%&{\rm ErrorRate}_{\rm PCA} = \int\limits_{-\infty}^{+\infty} {\rm d}u \frac{\left[ 1 - {\rm erf}\left(u - \sqrt{\rho m_v} \right) \right]}{2} P_ {\rm max}^{r-1}(u)\, ,
%\\
%&{\rm Overlap}_{\rm PCA} = \frac{{\rm ErrorRate} -1/r}{1-1/r}
%\end{eqnarray}
%\normalsize
%with
%\small
%\begin{eqnarray}
%P_ {\rm max}^{r-1}(x) &=& (r-1)\frac{\exp(-x^2/2)}{\sqrt{2 \pi}} \left(\frac{1 + {\rm erf}(x)}{2} \right)^{r-2}
%\end{eqnarray}
%\normalsize%
%Where ${\rm erf}(x) \in [-1;1]$ is defined as ${\rm erf}(x) = \sqrt{\frac{2}{\pi}} \int^x_0 \exp\left( \frac{-x^2}{2} \right)$. This result is derived in the appendix (\ref{AppendixA}). $r(\rho - m_v)$ is the mean-squared error achieved by PCA and is well-known in the theory of low-rank perturbations of random matrices \cite{10.2307/3481698}.

\section{Analysis for Mixture of Gaussians}
We analyze now the state evolution with the $f_v$ and $f_s$ functions corresponding to the GMM (\ref{fv_GMM}-\ref{fs_GMM}) for general number of clusters $r$. The difficulty here lies in the fact that there is in general no analytic expression for updating equations (\ref{SE_M_V}-\ref{SE_M_S}). We observe that the following form of the order parameters is conserved under the update of SE equations for the 
\begin{eqnarray}
M_s^t &=& \frac{I_r b_s^t}{r} + (1-b_s^t)\frac{J_r}{r^2}
\label{SpecificForm_M_s}
\\
M_v^t &=& {b_v^t}I_r + b_{v,J}^t \frac{J_r}{r},
\label{SpecificForm_M_v}
%\\
%\textrm{Where}&,&(b_s^t,b_v^t,b_{v,J}^t) \in [0;1]^3 \nonumber
\end{eqnarray}
where $(b_s^t,b_v^t,b_{v,J}^t) \in [0;1]^3$, and $I_r$ and $J_r$ are respectively the identity matrix and the $r \times r$ matrix filled with 1. 
%The stability of such value of the order parameters has been tested through runs of the AMP algorithm. 
Having $(b_s^t,b_v^t,b_{v,J}^t) = (1,1,0)$ would mean that we have achieved perfect reconstruction of the ground truth, while $(b_s^t,b_v^t) = (0,0)$ means that we are not able to extract any information from the matrix $X$ beyond the average of the $k$ clusters $V_k$.

Rewriting eqs. (\ref{SE_M_V}-\ref{SE_M_S}) using (\ref{fv_GMM}-\ref{fs_GMM}) and (\ref{SpecificForm_M_s}-\ref{SpecificForm_M_s}) we get the following SE equations for the GMM
\begin{equation}
b_v^t = \frac{b_s^t \rho}{\frac{r}{\alpha} + b_s^t \rho} \,,
\quad \quad
b_s^{t+1} = {\cal M}_r\left(b_v^t \rho r\right) \,,
\end{equation}
where
\small
\begin{equation}
{\cal M}_r\left( x \right)= \frac{1}{r - 1}\left[ r \int
\frac{e^{\frac{x}{r} + u_1 \sqrt{\frac{x}{r}}}}
{
e^{\frac{x}{r} + u_1 \sqrt{\frac{x}{r}}} + \sum\limits_{i = 2}^r  e^{ u_i \sqrt{\frac{x}{r}} }
}
\prod\limits_{i=1}^r {\cal D}u_i -1  \right] \,. \label{Function_Cal_M}
\end{equation}
\normalsize
%\left[ \frac{{\rm d} u_i}{\sqrt{2\pi}} e^{\frac{-u_i^2}{2}} \right]
Where the $u_i$ are Gaussian variables of mean 0 and unit variance. The equations close on $b_v$ and $b_s$ and do not depend on the term $b_{v,J}$. We can combine these to obtain a single update equation for the scalar variable $b_s^t$
\begin{eqnarray}
b_s^{t+1} &=& {\cal M}_r\left(b_s^t \frac{\rho^2 }{\frac{1}{\alpha} + \frac{\rho b_s^t}{r}}\right) \,. \label{SE_ANY_R}
\end{eqnarray}
%Notice that this equations is closed on $b_{v}^t$, which got eliminated. 
%{\bf Lenka: Is this true?}
%and we do not report the analogous equation for $b_{v}^t$, since these terms do not appear in the above equation.
The function ${\cal M}_r$, including the expansions for small $\rho$ and for large $r$, has been studied previously in \cite{lesieur2015mmse}. From a numerical point of view ${\cal M}_r(x)$ can be effectively computed with a Monte Carlo scheme if one takes advantage of the permutation symmetry of the Gaussian variables.

\subsection{Analysis of the phase transitions}

Note that $b_s={\cal M}_r(0) = 0$ is always a fixed point that we will call {\it uninformative}. By expanding ${\cal M}_r(x)$ around 0 one gets:
\begin{equation}
b_s^{t+1} = \frac{\alpha b_s^{t}\rho^2}{r^2} + \frac{\alpha^2 {b_s^t}^2}{2}\left[ r-4 - \frac{2r}{\rho}\right] \frac{\rho^4}{r^4} \,. \label{Expansion_SE_b_s}
\end{equation}
We are interested in when the uninformative fixed point becomes numerically unstable. From the expansion of (\ref{Expansion_SE_b_s}) one deduces that this occurs when
\begin{equation}
\rho > \rho_c = \frac{r}{\sqrt{\alpha}} \,.
\end{equation}
Looking at the second derivative of (\ref{Expansion_SE_b_s}), we deduce that when the uninformative fixed point becomes unstable, the second derivative is proportional to $r - 4 - 2 \sqrt{\alpha}$; if this is negative then this means that another stable fixed point appears close to 0 for $\rho > \rho_c$. If on the other hand the second derivative is positive when $\rho$ increases and crosses $\rho_c$, then the new stable fixed point will not be close to zero and we see a discontinuous jump in the MSE achieved by initializing the iteration close to the uninformative fixed point. This phenomenon is known as a first-order transition. 
%This means that for some value of $\rho,r$ and $\alpha$ there can be multiple stable fixed point to the SE equations.
If one fixes the number $r$ and $\alpha$, we have a first order transition in the GMM if $r > r_c = 4 + 2 \sqrt{\alpha}$.

\begin{figure}
\hspace{-0.5cm}
\includegraphics[scale=0.9]{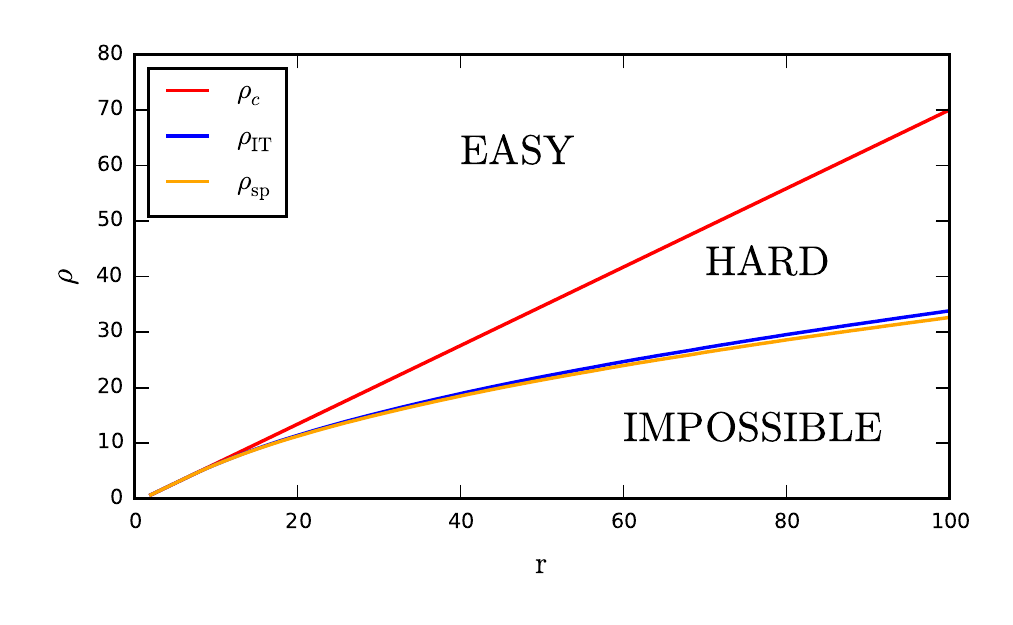}
\vspace{-1cm}
\caption{ The three phase transitions as a function of the number of clusters $r$ and the signal-to-noise parameter $\rho$, for sample-to-dimension ratio $\alpha=2$. In red is the algorithmic phase transition $\rho_c=r/\sqrt{\alpha}$, in blue the information theoretic threshold $\rho_{\rm IT}$ and in yellow the spinodal threshold $\rho_{\rm sp}$. As the number of clusters $r$ grows the gap between the algorithmic and the information-theoretic transition grows larger. }
\label{STATIC_SPINODAL_NORMALISED}
\end{figure}

It turns out that this sufficient criteria is also necessary. In order to analyze the SE numerically we introduce two ways to initialize the equations
\begin{itemize}
\item Uninformative initialization $b_s^{t=0} \approx 0$ : Initializing in such a way is equivalent to assuming that we use no knowledge about the ground truth signal. From an AMP point of view this means starting with means of messages close to 0. We denote $b_{\rm AMP}$ the reached fixed point. 
\item Informative initialization $b_s^{t=0} = 1$: This means initializing in the ground truth configuration. From an AMP point of view this corresponds to start with messages equal to the solution.  Of course, in reality the algorithm would not have access to the solution; but here we apply this initialization in order to track the corresponding fixed point. We denote $b_{\rm inf}$ the reached fixed point. 
\end{itemize}
For $r < 4 + 2\sqrt{\alpha}$ iterations with these two initializations lead to the same fixed point and the situation looks qualitatively as in Fig.~\ref{fig: BPQv} for two clusters. For $\rho< \rho_c$ we reach the uninformative fixed point $b_s=0$, whereas for $\rho> \rho_c$ we reach a fixed point with positive overlap with the ground truth, $b_s >0$. 

If $r > 4 + 2\sqrt{\alpha}$, then depending on the value of $\rho$ four different situations happen:
\begin{itemize}
\item $\rho>\rho_c$ : Here $b_{\rm AMP}=b_{\rm inf}>0$: The AMP algorithm is able to reach the information theoretically optimal reconstruction. This is called the {\bf EASY} phase.
\item $\rho_c>\rho>\rho_{\rm IT}$: Here $b_{\rm inf} > b_{\rm AMP}=0$, i.e. there are two fixed points to the SE equations and $b_{\rm inf}$ leads to a lower free energy (\ref{BetheFreeEnergy_UV_DE}). The AMP algorithm starting from the uninformative initialization is not able to reach the information theoretically optimal performance. Nevertheless it is possible, although in exponential time, to find a better fixed point with lower free energy. This is called  the {\bf HARD} phase. 
\item $\rho_{\rm IT} > \rho >\rho_{\rm sp}$: Here still $b_{\rm inf} > b_{\rm AMP}=0$, but now $b_{\rm AMP}$ has a lower free energy (\ref{BetheFreeEnergy_UV_DE}). The AMP algorithm starting from the informative state is able to reach a good fixed point correlated with the solution. However, finding this fixed point without prior knowledge of the solution is information theoretically impossible: it is hidden among an exponential number of other fixed points of AMP that have similar likelihood. This is called the information-theoretically {\bf IMPOSSIBLE} phase.
\item $\rho_{\rm sp} > \rho$: Here $b_{\rm AMP} = b_{\rm inf}=0$, i.e. there is only one fixed point to the SE equations. This is also the  information-theoretically {\bf IMPOSSIBLE} phase.
\end{itemize}

To compute these transitions $\rho_{\rm IT}$ and $\rho_{\rm sp}$ numerically we consider one value of $b_s = {\cal M}_r(x)$ and ask what is the value of $\rho$ such that $b_s = {\cal M}_r(x)$ is a fixed point. Using (\ref{SE_ANY_R}) the answer is
\begin{equation}
\rho(x,r) = \frac{x}{2r} + \sqrt{ \frac{x^2}{4r^2} + \frac{x}{\alpha {\cal M}_r(x)}} \label{RHO_X_R} \,.
\end{equation}
The spinodal transition $\rho_{\rm sp}$ is the minimum value of $\rho$ for which a fixed point other than 0 exists. We can estimate this by minimizing $\rho(x,r)$ (\ref{RHO_X_R}) with respect to $x$ and at fixed~$r$.
The  information-theoretic transition $\rho_{\rm IT}$ is obtained by expressing the difference in the free energy between $b_s = 0$ and ${\cal M}_r(x)$ at a given $\rho$ and then requiring this  quantity to be 0. It is possible to express the Bethe free energy using ${\cal M}_r$. If one integrates the gradient of $\phi_{\rm B}$ along the path $g(u)$ defined by
\begin{equation}
\forall u \in [0,{\cal M}_r(x)],g(u) = \left( u, u \frac{\rho(x)^2/r }{\frac{1}{\alpha} + \frac{\rho(x) u}{r}} \right) \,,
\end{equation}
after integrating by parts
\begin{eqnarray}
&&\phi_{\rm B}(0, \rho(x,\alpha,r),\alpha;r) \label{Zero_Energy} \\
&-&\phi_{\rm B}({\cal M}_r(x) , \rho(x,\alpha,r),\alpha,r) \nonumber \\
&=&\alpha \frac{r-1}{2 r^2}\!\!\left[ \int\limits_0^x \! {\rm d}u {\cal M}_r(u) +\int \limits_0^{{\cal M}_r(x)}\!\!\!\! \! \!{\rm d}u \! \frac{u \rho^2}{\frac{1}{\alpha} + \frac{u \rho(x)}{r}}   - x {\cal M}_r(x)\right]\!\!.\! \nonumber  
\end{eqnarray}
The  information-theoretic transition is found where $b_{\rm AMP}(r,\alpha,\rho)$ and $b_{\rm inf}(r,\alpha,\rho)$ both have the same free energy. The behavior of $\rho_c$, $\rho_{\rm IT}$ and $\rho_{\rm sp}$ as a function of $\rho$ for $\alpha=2$ is illustrated in Fig.~\ref{STATIC_SPINODAL_NORMALISED}. 

%Through the use of  (\ref{RHO_X_R},\ref{Zero_Energy}) one is able to find the position of both the static transition and the spinodal transition for any $r$ and $\alpha$.

\subsection{Large number of clusters}

Formulas (\ref{RHO_X_R}-\ref{Zero_Energy})  also allow us to explore the large $r$ limit of these solutions. From \cite{lesieur2015mmse} we know that
\begin{equation}
\forall \beta > 0,\lim_{r \rightarrow \infty} {\cal M}_r(\beta r \log r) = 1_{\beta > 2} \label{LIMIT_M_r} \,.
\end{equation}
We can compute the asymptote of $\rho_{\rm IT}$ and $\rho_{\rm sp}$ when $r \rightarrow \infty$.
We can do that by setting $x = \beta r \log(r)$ and then replacing ${\cal M}_r$ by (\ref{LIMIT_M_r}) in (\ref{RHO_X_R}) and (\ref{Zero_Energy}). 

To get $\rho_{\rm sp}$ one minimizes (\ref{RHO_X_R}) with respect to $\beta$. 
After some computation one gets 
\begin{equation}
\lim\limits_{r \rightarrow +\infty }\frac{\rho(\beta r \log(r),\alpha)}{\sqrt{r \log(r)}} =  \left\{ \begin{array}{ccc}
+\infty,\: {\rm if}\, \beta <  2 \\
\sqrt{\beta/\alpha},\: {\rm if}\, \beta \geq 2
\end{array} \right.
\end{equation}
Therefore in the large $r$ limit $\rho(\beta r \log(r))$ is minimized by taking  $\beta = 2$ and one gets
\begin{equation}
\rho_{\rm sp}(r,\alpha)= \sqrt{\frac{2 r \log r}{\alpha}}(1 + o_r(1)) \,.
\end{equation}

To get $\rho_{\rm IT}$ one finds the $\beta$ such that (\ref{Zero_Energy}) is set to 0.
In order to do that we write $x_{\rm IT}= \beta_{\rm IT} r \log(r)$. We take $\beta_{\rm IT} > 2$ since 
we want to have $\rho$ to be above $\rho_{\rm sp}$ so that there are multiple fixed point to the SE equations.
For $\beta > 2$ to leading order one has
\begin{equation}
\rho(x = \beta r \log(r),r) = \sqrt{\frac{\beta r \log(r)}{\alpha}}(1 + o_r(1))\,.
\end{equation}	
Therefore,
\begin{equation}
\frac{\rho(x = \beta r \log(r),r)}{r} \ll \frac{1}{\alpha} \,.
\end{equation}
Using this, equation (\ref{Zero_Energy}) can be further simplified. One gets
\begin{multline}
\phi_{\rm B}(0, \rho(x_{\rm IT},\alpha,r),\alpha;r) -
\phi_{\rm B}({\cal M}_r(x_{\rm IT}) , \rho(x_{\rm IT},\alpha,r),\alpha,r)  \\
=0 \approx \alpha \frac{r-1}{2 r^2}\!\!\left[ \int\limits_0^{x_{\rm IT}} \! {\rm d}u {\cal M}_r(u) - \frac{x_{\rm IT} {\cal M}_r(x_{\rm IT})}{2}\right]\!\!,\!  
\end{multline}
By setting $x_{\rm IT} = \beta_{\rm IT} r \log(r)$ in this equation and  taking the $r \rightarrow \infty$ limit one gets
\begin{equation}
0 = \int\limits_0^{\beta_{\rm IT}} \! {\rm d}u 1(u > 2) - \frac{ \beta_{\rm IT} 1( \beta_{\rm IT} > 2)}{2} \,.
\end{equation}
This is solved for $\beta_{\rm IT} = 4$, therefore one has
\begin{equation}
\rho_{\rm IT}(r,\alpha) = 2\sqrt{\frac{r \log(r)}{\alpha}}(1 + o_r(1)) \,.
\end{equation}
Thus, for large number of clusters $r$, the gap between the information-theoretic detectability threshold $\rho_{\rm IT}$ and the algorithmic threshold $\rho_c = r/\sqrt{\alpha}$ becomes large.

\subsection{Algorithmic comparison}

%\subsection{Numerical simulations}

\begin{figure}
	\centering
	\includegraphics[scale=0.7]{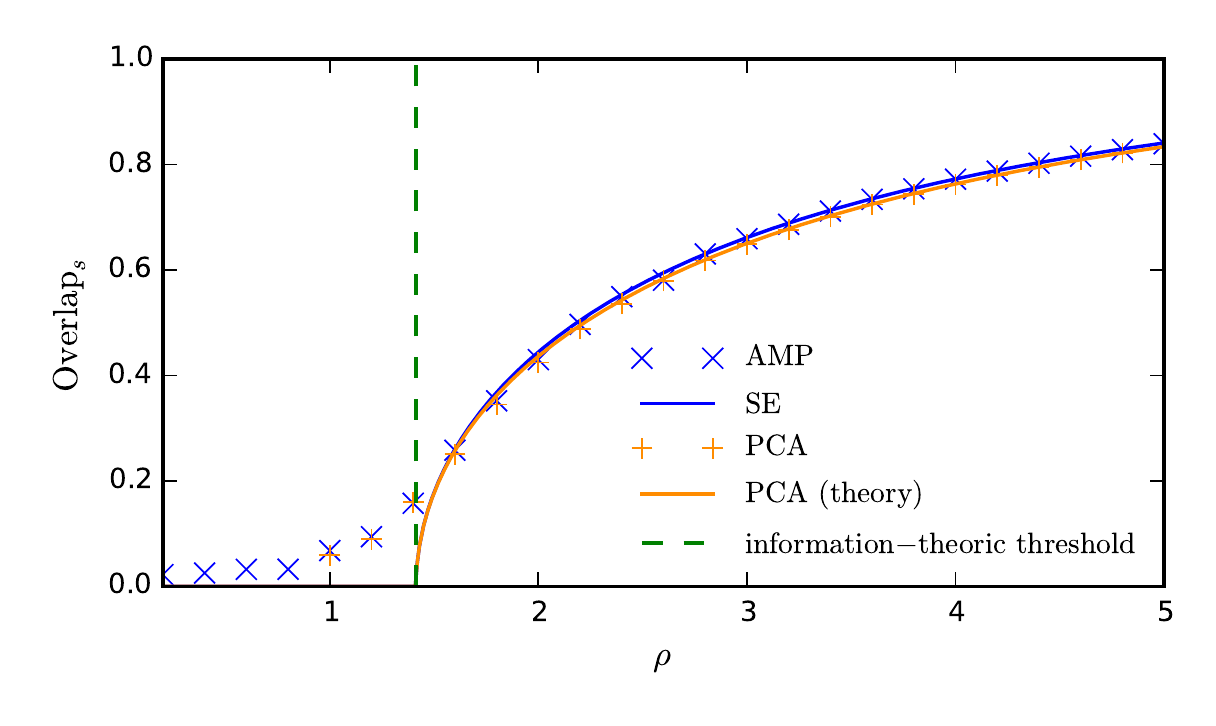} 
\vspace{-1cm}
	\caption{ Bayes-optimal overlap for clustering with $r=2$ clusters (full lines, using State Evolution), together with the results of numerical simulations (points), for $\alpha=2$, $n=1000$ and $m=2000$.  The overlap is defined in \eqref{eq:overlap}. The information-theoretic threshold at $\rho_c \approx 1.41$ is materialized by the green dashed line. Both AMP and PCA are able to perform better than chance beyond this transition. The performance of AMP and PCA are comparable in this case.}
\label{fig: BPQv}
\end{figure}

\begin{figure}
\centering
\includegraphics[scale=0.7]{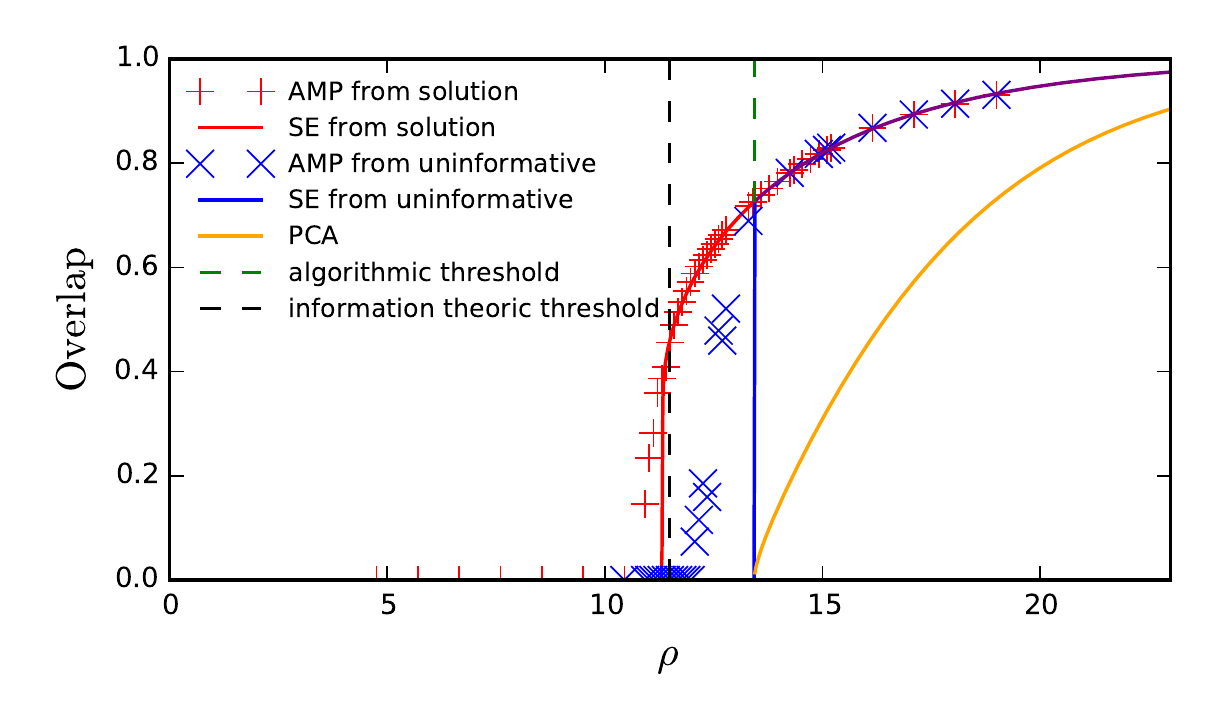}
\vspace{-1cm}
\caption{ Bayes-optimal overlap for clustering with $r=20$ clusters (full lines, using State Evolution), together with the results of numerical simulations (points), for $\alpha=2$, $n=10000$ and $m=20000$. In this case one observes an algorithmically hard phase, at $\rho_c \approx 14.1$ the overlap achieved by AMP in infinite size systems has a discontinuity. The algorithmic and information-theoretic thresholds are depicted by the vertical blue line and the dashed black line. Because of the discontinuous nature of the transition, the finite size effects are sizable in the hard region: the red points with non-zero overlap below $\rho_c$ show that AMP is able to reconstruct a fraction of the $r=20$ clusters. Increasing $n$ decreases these effects.}
\label{fig:Overlap_R20m_ALPHA2}
\end{figure}

%  ------- OVERLAP RANK=2 , all the algorithms  
Fig. \ref{fig: BPQv} and \ref{fig:Overlap_R20m_ALPHA2} contain numerical experiments with the AMP algorithm, and comparison with the theoretically predicted performance given by state evolution analysis. Data for both $r=2$ and $r=20$ clusters are presented. These two cases have qualitatively different properties. As predicted, the case with $r=20$ clusters exhibits a first order transition: there is a sharp jump in the overlap at $\rho_c$. Instead the case $r=2$ exhibits a second order transition: the overlap is continuous, only its derivative has a discontinuity. 

In both Fig. \ref{fig: BPQv} and \ref{fig:Overlap_R20m_ALPHA2} we also compare to the performance of the principal component analysis (PCA) performed on the matrix $X$. PCA is a standard spectral method to solve data clustering, one computes $r$ leading singular vectors of $X$ and instead of clustering $m$ points in $n$ dimensions, one concatenates the singular vectors into $m$ $r$-dimensional vectors and clusters in the $r$-dimensional space which is much simpler. The overlap reached with the PCA clustering follows from a more general theory of low-rank perturbations of random matrices \cite{10.2307/3481698}, but it can also be derived from the state evolution analysis of AMP as we present in the appendix. In this case of GMM with equal-size clusters the phase transition observed in PCA coincides with the phase transition of AMP $\rho_c$. Concerning the performance as measured by the overlap (\ref{eq:overlap}), we observe that although for two clusters the difference between the performance of the sub-optimal PCA and the Bayes-optimal AMP is hardy visible, for 20 clusters the performance of AMP close to the algorithmic transition $\rho_c$ is considerably better. 

%We can initialize the AMP equations in two ways. Either initializing using PCA or initializing in the solution. In the $r=2$ case the initialisation method used does not matter. But in the $r=20$ case these two methods will lead to different fixed point to both the SE equations and AMP equation. initializing using PCA or using the hidden solution is called either the uninformative or informative initialization. The difference between initialization is illustrated in Fig (\ref{fig:Overlap_R20m_ALPHA2}).


\section{Conclusion}

We analysed the problem of clustering high-dimensional data generated by the Gaussian mixture model.  We computed the asymptotic accuracy of the Bayes-optimal estimator and compared it to the accuracy obtained by the approximate message-passing algorithm. We located phase transitions in both the information-theoretic and algorithmic performance. 

%Although we have not done this here, we claim our analysis of state evolution can be made mathematically rigorous using the techniques of~\cite{DBLP:journals/corr/abs-1112-0708,DBLP:journals/corr/DeshpandeM14}.

Our main result is that, when the number of clusters is sufficiently large, eq.~(\ref{CriteriaFirstOrder}), there is a gap between the information-theoretic threshold---where it becomes possible to label points and find the cluster centers better than chance---and the computational threshold, at which polynomial-time algorithms such as PCA or AMP succeed.  This suggests that, as has been conjectured for analogous problems in statistical inference, there is a hard-but-detectable regime where clustering is information-theoretically possible, but computationally intractable.

%This particuliar setting we analysed allows us to illustrate the true force of these AMP algorithms, which is that, provided one manages to make them converge, really shine when dealing with system that have hard-constraints (such as sparsity or belonging to a single group).

\section*{Acknowledgments}
CM, FK and LZ  thank the Simons Institute in Berkeley for its hospitality. FK acknowledge funding from the EU (FP/2007-2013/ERC grant agreement
307087-SPARCS).

\begin{appendices}
\section{Performance of PCA}
\label{AppendixA}

Using state evolution one can analyze the distribution of the top eigenvectors of $X$ and therefore the performance of PCA on the Gaussian mixture clustering problem. Let us write the density evolution equation the posterior measure (\ref{DensityProb}) with $X$ created using a GMM, and where we have replaced the Bayes-optimal prior on $v_i$ and $s_j$ with Gaussian priors
\begin{equation}
P_v(v_i) = \frac{\exp\left( \frac{- \Vert v_i \Vert_2^2}{2} \right)}{\sqrt{2 \pi}^r},\: P_s(s_j) = \frac{\exp\left( \frac{- r \Vert s_j \Vert_2^2}{2} \right)}{\sqrt{2 \pi / r}^r}\, .  \label{G_prior}
\end{equation}
Let us define $V_{\rm PCA}$, $S_{\rm PCA}$ the matrices containing the first $r$ singular vectors of $X$. Where $V_{\rm PCA} \in \mathbb{R}^{n \times r}$, $S_{\rm PCA} \in \mathbb{R}^{m \times r}$. $S_{\rm PCA}$ defines $m$ points in a $r$ dimensional space. We expect the $r$ dimensional points extracted from $S_{\rm PCA}$ to form a mixture of Gaussians. Let us write the AMP for Gaussian priors.  The general AMP equations for the mismatching priors can be found in \cite{NIPS2013_5074}. One gets
\begin{eqnarray}
\hat{V}^{t} &=& \left( \sqrt{\frac{\rho}{n}} X \hat S^t  - \rho  \alpha \hat V^{t-1} \Sigma_s^t \right)\Sigma_v^{t} \,,
\\
\Sigma_v^{t} &=& \left(I_r + \rho/n {\hat{S}^{t^\top} S^t}\right)^{-1} \,,
\label{Gaussian_V_Sigma}
\\
\hat{S}^{t+1} &=& \left( \sqrt{\frac{\rho}{n}} X^\top \hat V^t  - \rho \hat S^{t} \Sigma_v^t \right)\Sigma_s^{t+1} \label{S_update} \,,
\\
\Sigma_s^{t+1} &=& \left(r I_r + \rho/n {\hat{V}^{t^\top} V^t}\right)^{-1} \,.
\label{Gaussian_S_Sigma}
\end{eqnarray}
We aim to prove that at a fixed point $\hat V$ and $\hat S$ will be singular values of $X$.

After simplifications of the above equations one gets
\begin{eqnarray}
\hat{V} &=& \sqrt{\frac{\rho}{n}} X \hat{S} \left(\Sigma_v^{-1} + \rho \alpha \Sigma_s \right)^{-1}
\label{Eq_Eig_FixedPoint_V} \,,
\\
\hat{S} &=& \sqrt{\frac{\rho}{n}} X^\top \hat{V} \left(\Sigma_s^{-1} + \rho \Sigma_v \right)^{-1}
\label{Eq_Eig_FixedPoint_S} \,.
\end{eqnarray}
By putting this in (\ref{Eq_Eig_FixedPoint_V}) in (\ref{Eq_Eig_FixedPoint_S}) one gets
\begin{equation}
\hat{S} = \frac{\rho}{n} X^\top X \hat{S} \left(\Sigma_v^{-1} + \rho \alpha \Sigma_s \right)^{-1} \left(\Sigma_s^{-1} + \rho \Sigma_v \right)^{-1} \,.
\label{Equation_Eigen_S}
\end{equation}
This means that the columns of $\hat{S}$ are a linear combination of eigenvectors of $X^\top X$ that is right eigenvector of $S$.

We can also write the corresponding state evolution equations.
One gets
\small
\begin{equation}
M_v^{t} = \mathbb{E}_{W,v_0}\left[ f_v\left(\alpha \rho M_s^t , \alpha \rho M_s^t v_0 + \sqrt{\alpha \rho M_s^t} W\right)v_0^\top \right] 
\,, \nonumber
\end{equation}
\begin{equation}
Q_v^{t} = \mathbb{E}_{W,v_0}\left[ f_v\left(\alpha \rho M_s^t , \alpha \rho M_s^t v_0 + \sqrt{\alpha \rho M_s^t} W\right)f_v(\cdots,\cdots) \right] 
\,,\nonumber
\end{equation}
\begin{equation}
M_s^{t+1} = \mathbb{E}_{W,s_0}\left[ f_s\left( \rho M_v^t , \rho M_v^t s_0 + \sqrt{\rho M_v^t} W\right)s_0^\top \right]
\,,\nonumber
\end{equation}
\begin{equation}
Q_s^{t+1} = \mathbb{E}_{W,s_0}\left[ f_s\left( \rho M_v^t , \rho M_v^t s_0 + \sqrt{\rho M_v^t} W\right)  f_s(\cdots,\cdots) \right]
\,,\nonumber
\end{equation}
\normalsize
With the Gaussian priors (\ref{G_prior}) this gives us
\begin{eqnarray}
M_v^{t} &=& (I_r + \alpha \rho Q_s^t)^{-1} \alpha \rho M_s^t \,,
\\
Q_v^{t} &=& (I_r + \alpha \rho Q_s^t)^{-2} \left( \alpha^2 \rho^2 M_s^t + \alpha \rho Q_s^t\right) \,,
\\
M_s^{t} &=& (r I_r + \rho Q_v^t)^{-1} \rho M_v^t/r \,,
\\
Q_s^{t} &=& (r I_r + \rho Q_v^t)^{-2} \left(\rho^2 M_v^t /r+ \rho Q_v^t\right) \,.
\end{eqnarray}
Experiments with these equations show that stable fixed point are always of the form
\begin{eqnarray}
M_v^t &=& m_v^t R, \quad Q_v^t = m_v^t I_r\, ,
\label{Gaussian_Shape_Mv}
\\
M_s^t &=& \frac{m_s^t R}{r}, \quad Q_s^t = \frac{m_s^t I_r}{r}\, ,
\label{Gaussian_Shape_Ms}
\end{eqnarray}
where $R$ is some rotation matrix. The new update equation become
\begin{eqnarray}
m_v^{t+1} &=& \frac{\rho \alpha m_s^t}{1 + \rho \alpha m_s^t}\, ,
\\
m_s^{t+1} &=& \frac{\rho m_v^t}{r + \rho m_v^t}\, .
\end{eqnarray}
Let us write
\begin{eqnarray}
m_v &=& m_v^{t\to \infty} = \max\left( \frac{\alpha \rho^2 - r^2}{r + \rho \alpha} , 0 \right)\, ,
\\
m_s &=& m_s^{t \to \infty} = \max\left( \frac{\alpha \rho^2 - r^2}{r \rho \alpha + \rho^2 \alpha} , 0 \right)\, ,
\end{eqnarray}

By combining (\ref{Gaussian_S_Sigma}), (\ref{Gaussian_V_Sigma}), (\ref{Gaussian_Shape_Mv}) and (\ref{Gaussian_Shape_Ms}) one sees that in the large $n$ limit $\Sigma_v$ and $\Sigma_s$ are proportional to the identity. Therefore using (\ref{Equation_Eigen_S}) one gets
\begin{eqnarray}
\hat{S} a &=& {\frac{\rho}{n}} X^\top X \hat{S} \,, \label{Equation_Eigen_V_2}
\end{eqnarray}
where $a$ is some number.
This means that $\hat{S}$ is proportional to the first eigenvectors. But the AMP analysis also tells us how $\hat{S}$ will be distributed. The details of the analysis of state evolution can be found in \cite{lesieur2015mmse} in the $XX^\top$ case
\begin{equation}
\hat{s}_j = f_s\left( \rho m_v I_r , \rho m_v R s_{0,j} + \sqrt{\rho m_v} W\right)\, .
\end{equation}
Where $W$ is a Gaussian variable given by ${\cal N}(0,I_r)$.
Up to a proportionality constant one has.
\begin{equation}
\hat{s}_j = \sqrt{ \rho m_v }R s_{0,j} + W\, .
\end{equation}
This means that the first $r$ right eigenvectors of $R$ are distributed as a mixture of Gaussians in a $r$ dimensional space where the centers are placed at positions $\sqrt{\rho m_v} R e_k$ and the noise is a Gaussian white noise of zero mean and covariance matrix $I_r$. The questions now becomes how well can one cluster this mixture of Gaussians? Since we are dealing with a finite-dimensional space $r=O(1)$ and a large number of points $m \rightarrow \infty$ then we know that we should be able to learn the parameters from this mixture of Gaussian perfectly.

The problem now becomes given a point $\hat{v}_j$  what is the chance that it was created using point $v_{0,j}$ this is done by maximizing the likelihood. To find back with what $e_k$ $\hat{v}_j$ was created one needs to compute.
\small
\begin{equation}
\hat{k}(\hat{v_j}) = {\rm argmax} \left\{ k \in [1;r], \frac{1}{r \sqrt{2 \pi}} \exp\left( \frac{-\Vert \hat{v}_j - \sqrt{\rho m_v} \vec{e}_k  \Vert^2_2}{2} \right)   \right\}\nonumber
\end{equation}
\normalsize
or
%\small
\begin{equation}
\hat{k}(\hat{v_j}) = {\rm argmax} \left\{ k \in [1;r],  \langle \hat{v}_j ; \vec{e}_k \rangle \right\}\,.
\end{equation}
%\normalsize
One need to compute with what probability $\hat{k}$ is the right $k$
\begin{multline}
P( \hat{k}(   \sqrt{\rho m_v} \vec{e}_{k_0} + W ) = k_0) = \\ P( W_{k_0} + \sqrt{\rho m_v} > \max\limits_{k \in [1;r],k \neq k_0}\left\{ W_k  \right\} )\, . \nonumber
\end{multline}
This can be computed using the distribution of the maxima of random variables of zero mean and variance $r-1$ Gaussian variables.
\begin{multline}
P( \hat{k}(   \sqrt{\rho m_v} \vec{e}_{k_0} + W ) = k_0) = \\ \int\limits_{-\infty}^{+\infty} {\rm d}u\left[ 1/2 - 1/2 {\rm erf}\left(u - \sqrt{\rho m_vb } \right) \right] P_ {\rm max}^{r-1}(u)\, ,
\end{multline}
where $P_ {\rm max}^{r-1}$ is the density probability of the maximum of $r-1$ independent Gaussian variables of mean 0 and unit variance.
Where ${\rm erf}(x) \in [-1;1]$ is defined as ${\rm erf}(x) = \sqrt{\frac{2}{\pi}} \int^x_0 \exp\left( \frac{-x^2}{2} \right)$. 
\begin{equation}
P_{\rm max}^{r-1}(u) = \frac{r-1}{\sqrt{2 \pi}} \exp( -u^2/2) \left[\frac{1 + {\rm erf}\left( u\right)}{2} \right]^{r-2}\, .
\end{equation}
Therefore one can compute the average number of errors made using PCA to cluster the data points.
\begin{equation}
{\rm ErrorRate}_{\rm PCA} = \int\limits_{-\infty}^{+\infty} {\rm d}u \frac{\left[ 1 - {\rm erf}\left(u - \sqrt{\rho m_v} \right) \right]}{2} P_ {\rm max}^{r-1}(u)\, . \nonumber
\end{equation}
The mean-squared error achieved by PCA is then $r(\rho - m_v)$ and is known in the theory of low-rank perturbations of random matrices \cite{10.2307/3481698}.

\end{appendices}

\bibliographystyle{IEEEtran}
\bibliography{refs}

\begin{thebibliography}{10}
\providecommand{\url}[1]{#1}
\csname url@samestyle\endcsname
\providecommand{\newblock}{\relax}
\providecommand{\bibinfo}[2]{#2}
\providecommand{\BIBentrySTDinterwordspacing}{\spaceskip=0pt\relax}
\providecommand{\BIBentryALTinterwordstretchfactor}{4}
\providecommand{\BIBentryALTinterwordspacing}{\spaceskip=\fontdimen2\font plus
\BIBentryALTinterwordstretchfactor\fontdimen3\font minus
  \fontdimen4\font\relax}
\providecommand{\BIBforeignlanguage}[2]{{%
\expandafter\ifx\csname l@#1\endcsname\relax
\typeout{** WARNING: IEEEtran.bst: No hyphenation pattern has been}%
\typeout{** loaded for the language `#1'. Using the pattern for}%
\typeout{** the default language instead.}%
\else
\language=\csname l@#1\endcsname
\fi
#2}}
\providecommand{\BIBdecl}{\relax}
\BIBdecl

\bibitem{hoyle2004principal}
D.~C. Hoyle and M.~Rattray, ``Principal-component-analysis eigenvalue spectra
  from data with symmetry-breaking structure,'' \emph{Physical Review E},
  vol.~69, no.~2, p. 026124, 2004.

\bibitem{10.2307/3481698}
J.~Baik, G.~B. Arous, and S.~P\'ech\'e, ``Phase transition of the largest
  eigenvalue for nonnull complex sample covariance matrices,'' \emph{The Annals
  of Probability}, vol.~33, no.~5, pp. 1643--1697, 2005.

\bibitem{mezard2009information}
M.~M{\'e}zard and A.~Montanari, \emph{Information, Physics, and Computation},
  ser. Oxford Graduate Texts.\hskip 1em plus 0.5em minus 0.4em\relax OUP
  Oxford, 2009.

\bibitem{rangan2012iterative}
S.~Rangan and A.~K. Fletcher, ``Iterative estimation of constrained rank-one
  matrices in noise,'' in \emph{IEEE International Symposium on Information
  Theory Proceedings (ISIT)}.\hskip 1em plus 0.5em minus 0.4em\relax IEEE,
  2012, pp. 1246--1250.

\bibitem{NIPS2013_5074}
R.~Matsushita and T.~Tanaka, ``Low-rank matrix reconstruction and clustering
  via approximate message passing,'' in \emph{Advances in Neural Information
  Processing Systems 26}, 2013, pp. 917--925.

\bibitem{javanmard2013state}
A.~Javanmard and A.~Montanari, ``State evolution for general approximate
  message passing algorithms, with applications to spatial coupling,''
  \emph{Information and Inference}, 2013.

\bibitem{krzakala2016mutual}
F.~Krzakala, J.~Xu, and L.~Zdeborov{\'a}, ``Mutual information in rank-one
  matrix estimation,'' \emph{ITW 2016, arXiv:1603.08447}, 2016.

\bibitem{DBLP:journals/corr/BarbierDMKLZ16}
J.~Barbier, M.~Dia, N.~Macris, F.~Krzakala, T.~Lesieur, and L.~Zdeborov{\'{a}},
  ``Mutual information for symmetric rank-one matrix estimation: {A} proof of
  the replica formula,'' \emph{NIPS 2016, arxiv:1606.04142}, 2016.

\bibitem{lloyd1982least}
S.~P. Lloyd, ``Least squares quantization in {PCM},'' \emph{IEEE Transactions
  on Information Theory}, vol.~28, no.~2, pp. 129--137, 1982.

\bibitem{cover2012elements}
T.~M. Cover and J.~A. Thomas, \emph{Elements of information theory}.\hskip 1em
  plus 0.5em minus 0.4em\relax John Wiley \& Sons, 2012.

\bibitem{watkin1994optimal}
T.~Watkin and J.-P. Nadal, ``Optimal unsupervised learning,'' \emph{Journal of
  Physics A: Mathematical and General}, vol.~27, no.~6, p. 1899, 1994.

\bibitem{barkai1994statistical}
N.~Barkai and H.~Sompolinsky, ``Statistical mechanics of the maximum-likelihood
  density estimation,'' \emph{Physical Review E}, vol.~50, no.~3, p. 1766,
  1994.

\bibitem{biehl1994statistical}
M.~Biehl and A.~Mietzner, ``Statistical mechanics of unsupervised structure
  recognition,'' \emph{Journal of Physics A: Mathematical and General},
  vol.~27, no.~6, p. 1885, 1994.

\bibitem{buhot1998phase}
A.~Buhot and M.~B. Gordon, ``Phase transitions in optimal unsupervised
  learning,'' \emph{Physical Review E}, vol.~57, no.~3, p. 3326, 1998.

\bibitem{MezardParisi87b}
M.~M{\'e}zard, G.~Parisi, and M.~A. Virasoro, \emph{Spin-Glass Theory and
  Beyond}, ser. Lecture Notes in Physics.\hskip 1em plus 0.5em minus
  0.4em\relax Singapore: World Scientific, 1987, vol.~9.

\bibitem{zdeborova2015statistical}
L.~Zdeborová and F.~Krzakala, ``Statistical physics of inference: thresholds
  and algorithms,'' \emph{Advances in Physics}, vol.~65, no.~5, pp. 453--552,
  2016.

\bibitem{thouless1977solution}
D.~J. Thouless, P.~W. Anderson, and R.~G. Palmer, ``Solution of'solvable model
  of a spin glass','' \emph{Philosophical Magazine}, vol.~35, no.~3, pp.
  593--601, 1977.

\bibitem{bayati2011dynamics}
M.~Bayati and A.~Montanari, ``The dynamics of message passing on dense graphs,
  with applications to compressed sensing,'' \emph{IEEE Transactions on
  Information Theory}, vol.~57, no.~2, pp. 764--785, 2011.

\bibitem{bayati2015universality}
M.~Bayati, M.~Lelarge, A.~Montanari \emph{et~al.}, ``Universality in polytope
  phase transitions and message passing algorithms,'' \emph{The Annals of
  Applied Probability}, vol.~25, no.~2, pp. 753--822, 2015.

\bibitem{DBLP:journals/corr/DeshpandeM14}
Y.~Deshpande and A.~Montanari, ``Information-theoretically optimal sparse
  {PCA},'' in \emph{Information Theory (ISIT), 2014 IEEE International
  Symposium on}.\hskip 1em plus 0.5em minus 0.4em\relax IEEE, 2014, pp.
  2197--2201.

\bibitem{deshpande2015asymptotic}
Y.~Deshpande, E.~Abbe, and A.~Montanari, ``Asymptotic mutual information for
  the binary stochastic block model,'' in \emph{2016 IEEE International
  Symposium on Information Theory (ISIT)}.\hskip 1em plus 0.5em minus
  0.4em\relax IEEE, 2016, pp. 185--189.

\bibitem{lesieur2015mmse}
T.~Lesieur, F.~Krzakala, and L.~Zdeborov\'a, ``Mmse of probabilistic low-rank
  matrix estimation: Universality with respect to the output channel,'' in
  \emph{2015 53rd Annual Allerton Conference on Communication, Control, and
  Computing (Allerton)}, 2015.

\bibitem{KoradaMacris}
S.~Korada and N.~Macris, ``Exact solution of the gauge symmetric p-spin glass
  model on a complete graph,'' \emph{Journal of Statistical Physics}, vol. 136,
  no.~2, pp. 205--230, 2009.

\bibitem{banks2016information}
J.~Banks, C.~Moore, R.~Vershynin, and J.~Xu, ``Information-theoretic bounds and
  phase transitions in clustering, sparse pca, and submatrix localization,''
  \emph{arXiv preprint arXiv:1607.05222}, 2016.

\end{thebibliography}
\end{document}